\documentclass[11pt,a4paper]{article}
\usepackage[a4paper,margin=0.95in]{geometry}
\usepackage[T1]{fontenc}
\usepackage{lmodern}
\usepackage{graphicx} 
\usepackage{booktabs} 
\usepackage{multirow} 
\usepackage{array} 
\usepackage{float} 
\usepackage{subcaption} 
\usepackage{amsmath} 
\usepackage{amssymb} 
\usepackage{xcolor} 
\usepackage{listings} 
\usepackage{caption} 
\usepackage{hyperref} 
\usepackage{enumitem}
\usepackage{parskip}
\usepackage{titling}
\usepackage{xurl}
\usepackage{makecell}
\usepackage{biblatex}
\definecolor{reportgreen}{HTML}{048012}
\definecolor{yamlkey}{RGB}{0,70,140} 
\definecolor{yamlstring}{RGB}{160,40,40} 
\definecolor{yamlcomment}{RGB}{80,120,80} 
\definecolor{yamlbackground}{RGB}{248,248,248}
\lstdefinelanguage{yaml}{ 
    keywords={true,false,null}, 
    keywordstyle=\color{blue}\bfseries, 
    basicstyle=\ttfamily\footnotesize, 
    sensitive=false, comment=[l]{\#}, 
    commentstyle=\color{yamlcomment}\itshape, 
    stringstyle=\color{yamlstring}, 
    morestring=[b]', 
    morestring=[b]" 
} 
\lstdefinestyle{yamlstyle}{ 
    language=yaml, 
    basicstyle=\ttfamily\footnotesize, 
    backgroundcolor=\color{yamlbackground}, 
    frame=single, 
    rulecolor=\color{gray}, 
    breaklines=true, 
    breakatwhitespace=false, 
    columns=fullflexible, 
    keepspaces=true, 
    showstringspaces=false, 
    tabsize=2, 
    numbers=left, 
    numberstyle=\tiny\color{gray}, 
    numbersep=8pt, 
    xleftmargin=1.5em, 
    framexleftmargin=1em 
}

\hypersetup{
colorlinks=true,
linkcolor=blue,
citecolor=blue,
urlcolor=blue
}
\title{\textbf{Beneath the Tokens: A Performance Engineering Study of Multi-Token Prediction in GPU-Accelerated LLM Inference}}
\author{Suwesh Prasad Sah\\
\small \href{https://suwesh.github.io}{suwesh.github.io}}

\date{September 2026}
\begin{document}

\maketitle
\begin{abstract}
    Autoregressive large language model inference repeatedly invokes the target model to generate one token at a time, making generation sensitive to GPU memory movement and sequential execution. This study evaluates two-token multi-token prediction (MTP) against autoregressive decoding in a controlled single-request deployment on an NVIDIA A10G GPU. A 360-request benchmark covered plain-text, reasoning-intensive, and tool-calling workloads, while runtime telemetry, Nsight Systems, PyTorch Profiler, and selected Nsight Compute measurements were used to explain the observed performance.
    
    MTP increased output throughput by \(1.91\times\) to \(2.19\times\) across all prompts and reduced time to first output by 10.0--14.2\%. Median mean acceptance length ranged from 2.370 to 2.595 tokens per verification iteration. Profiling showed that MTP introduced a longer and more complex execution path, including proposal, sampling, attention, gathering, and reduction operations. However, it required 56.4--78.1\% fewer executions of the selected repeating CUDA Graph per generated token. The dominant MTP GEMM kernel was not faster than the dominant autoregressive GEMV kernel, and selected instances of both approached the A10G memory-bandwidth limit. These results show that MTP improved inference through amortization: greater token progress reduced repeated GPU execution sufficiently to outweigh the additional speculative-execution cost.
\end{abstract}
\\
\noindent\textbf{Artifacts and Supporting Evidence:} \\ \href{https://github.com/suwesh/performance-engineering-series/tree/main/speculative-decoding-gpu-profiling}{Code and documentation} \enspace|\enspace \href{https://www.kaggle.com/datasets/suwesh/beneath-the-tokens-gpu-profiling}{Complete benchmark and profiling artifact archive}
\\
\textbf{Notice:} This manuscript is a technical report and has not undergone peer review.
\\
\noindent\textcolor{reportgreen}{\rule{\textwidth}{0.7pt}}
\section{Introduction}
Large language model (LLM) inference commonly generates output through autoregressive decoding, where each new token is mathematically conditioned on the entire preceding sequence. In practice, this sequential dependency forces the inference engine into a loop of single-token forward passes, which are severely bottlenecked by GPU memory bandwidth. Speculative decoding mitigates this hardware bottleneck by verifying multiple candidate tokens in parallel during a single target model execution step, potentially advancing the generation prefix by multiple tokens at once. Multi-token prediction (MTP) provides model-native predictions of multiple future tokens that can be used to support speculative inference~\cite{gloeckle2024better}. However, its practical benefit cannot be determined from the decoding algorithm alone. Additional speculative work is introduced, and the resulting wall-clock performance depends on whether greater token progress offsets the cost of that work.

This study evaluates two-token MTP in a controlled, GPU-accelerated LLM inference deployment. Autoregressive and MTP decoding are compared across plain-text, reasoning-intensive, and tool-calling workloads using a frozen experimental protocol. The evaluation combines clean application-level measurements with runtime acceptance metrics and matched profiler captures. This separation preserves the benchmark as the source of quantitative performance results while using profiling evidence only to investigate the execution behavior associated with those results.

The investigation focuses on three measurable aspects of speculative execution. First, MTP runtime metrics characterize the amount of useful progress obtained from speculative proposals. Second, matched Nsight Systems reports are used to compare repeated GPU execution cadence and kernel composition between autoregressive and MTP decoding. Third, matched PyTorch Profiler traces provide higher-level operational context for differences observed in the GPU timelines. Together, these evidence sources examine how speculative progress is translated into application-level latency and throughput without assuming that the observed acceleration results from individual kernels executing faster.

Accordingly, this paper addresses four research questions:
\begin{itemize}
    \item \textbf{RQ1}: How does two-token MTP affect time to first output, end-to-end latency, output throughput, and semantic-phase-associated duration across plain-text, reasoning-intensive, and tool-calling workloads?
    \item \textbf{RQ2}: How does observed speculative acceptance behavior correspond to the realized performance differences across workloads?
    \item \textbf{RQ3}: How does MTP change repeated GPU execution cadence and kernel composition relative to autoregressive decoding for comparable generated outputs?
    \item \textbf{RQ4}: Which observable runtime and GPU execution differences help explain the gap between speculative token progress and realized wall-clock acceleration?
\end{itemize}

The study contributes a controlled workload-level comparison of autoregressive and MTP decoding, a theory-guided evaluation of the relationship between speculative progress and realized performance, and a cross-layer analysis connecting application-level measurements with runtime and GPU execution evidence.

\section{Speculative Decoding}
\label{sec:speculative-decoding}
Speculative decoding was introduced as a method for accelerating autoregressive generation while preserving the output distribution of the target model~\cite{leviathan2023fast}. The method uses two models: a target model \(M_p\), whose output distribution must be preserved, and a less expensive draft model \(M_q\), which proposes candidate tokens. Rather than invoking the target model once for every generated token, the draft model first generates a sequence of \(\gamma\) candidate tokens. The target model then evaluates the corresponding candidate prefixes together and determines how many proposed tokens can be accepted.

Let \(p_i(x)\) and \(q_i(x)\) denote the token distributions produced by the target and draft models, respectively, at speculative position \(i\), after applying the configured sampling policy. A proposed token \(x_i \sim q_i\) is accepted with probability \begin{equation}
    a_i(x_i) = \min\left(1, \frac{p_i(x_i)}{q_i(x_i)}\right).
    \label{eq:acceptance-probability}
\end{equation}

where \(a_i(x_i)\) denotes the acceptance probability of the proposed token. If the target model assigns the proposed token at least as much probability as the draft model, then \(p_i(x_i) \geq q_i(x_i)\) and the proposal is always accepted. Otherwise, the proposal is accepted with probability \(p_i(x_i)/q_i(x_i)\).

The target distributions for the proposed positions are evaluated in parallel, but the resulting proposals are accepted in prefix order. If proposal \(x_i\) is rejected, all later proposals from the same speculative sequence are discarded because their conditioning prefix is no longer valid. A replacement token is then sampled from the corrected distribution \begin{equation}
    p_i'(x)
    =
    \operatorname{norm}
    \left(\max\left(0,p_i(x)-q_i(x)\right)\right).
    \label{eq:corrected-distribution}
\end{equation}

If all \(\gamma\) draft tokens are accepted, one additional token is sampled from the target model. This acceptance and correction procedure ensures that the resulting sequence follows the target model's output distribution, even though some tokens were initially proposed by the draft model~\cite{leviathan2023fast}.

Consequently, one speculative step emits between one and \(\gamma+1\) tokens. For example, if six tokens are proposed and all six are accepted, the step may emit the six accepted proposals and one additional target-model token. If only the first proposal is accepted, the step emits that accepted proposal followed by a target-derived replacement token. In the worst case, if no proposal is accepted, the step still emits one target-derived token. Speculative decoding can therefore reduce the number of sequential target-model steps when the draft model proposes tokens that are frequently accepted.

Let \(\alpha\) denote the mean probability that a draft token is accepted. Under the simplifying assumption that draft-token acceptance events are independent and identically distributed, the expected number of tokens emitted by one speculative step is \begin{equation}
    \mathbb{E}[N]
    =
    1+\alpha+\alpha^2+\cdots+\alpha^\gamma
    =
    \frac{1-\alpha^{\gamma+1}}{1-\alpha}.
    \label{eq:expected-token-progress}
\end{equation}

The leading term represents the token supplied by the target model, while each subsequent term represents the probability of accepting a progressively longer prefix of draft tokens. In the configuration evaluated in this study, \(\gamma=2\), giving \begin{equation}
    \mathbb{E}[N]
    =
    1+\alpha+\alpha^2,
    \qquad
    1 \leq \mathbb{E}[N] \leq 3.
    \label{eq:two-token-progress}
\end{equation}

Higher acceptance can therefore reduce the number of sequential verification steps required to produce an output sequence. This reduction does not guarantee an equal wall-clock speedup, because generating the proposals and verifying multiple positions introduce additional work. Let \(c\) denote the cost of one draft-model step relative to one target-model step. Under the assumptions of the original analytical model, the expected wall-clock improvement is \begin{equation}
    S_{\mathrm{theory}}
    =
    \frac{1-\alpha^{\gamma+1}}{(1-\alpha)(1+\gamma c)}
    =
    \frac{\mathbb{E}[N]}{1+\gamma c}.
    \label{eq:theoretical-speedup}
\end{equation}

Equation~\ref{eq:theoretical-speedup} expresses the central trade-off: speculative decoding is beneficial when the additional token progress obtained through accepted proposals outweighs the cost of producing those proposals. The coefficient \(c\) is not solely a property of the models; it also depends on the hardware and software implementation. The analytical model therefore provides a prediction about the balance between acceptance and cost, rather than guaranteeing a fixed speedup for every deployment. The original work likewise conditions its wall-time analysis on assumptions about draft cost and available computational concurrency~\cite{leviathan2023fast}.

The present study evaluates this trade-off for a two-token MTP configuration. In this deployment, a dedicated MTP drafter model supplies the speculative candidates, while the primary autoregressive target model performs verification and acceptance within the inference runtime. Runtime acceptance metrics characterize the useful progress obtained from speculation, and matched profiler captures examine the repeated GPU execution associated with obtaining that progress. The analysis therefore investigates whether the observed wall-clock acceleration is consistent with the theoretical balance between increased token progress and additional speculative-execution cost.

\section{Experimental Methodology}
\label{sec:methodology}
The study compares autoregressive decoding with two-token MTP speculative decoding under a controlled, single-request inference configuration. The experiment separates clean application-level benchmarking from profiler-instrumented executions. Benchmark runs provide the quantitative performance results, while runtime logs, Nsight Systems reports, and PyTorch Profiler traces provide supporting evidence about speculative acceptance and GPU execution behavior. All conditions were governed by a frozen experimental protocol whose integrity was verified using a SHA-256 checksum.

\subsection{System Under Test} 
\label{subsec:system-under-test} 
The experiments were conducted on an AWS EC2 instance equipped with one NVIDIA A10G GPU with 24~GB of device memory~\cite{nvidia2022a10g}. The host ran Red Hat Enterprise Linux 9.8 with Linux kernel 5.14.0-687.45.1.el9\_8.x86\_64. The captured environment reported NVIDIA driver version 595.71.05 and CUDA version 13.2. Inference was served through vLLM 0.24.0~\cite{kwon2023efficient} using its OpenAI-compatible streaming API. The experimental client and orchestration tools were implemented in Python 3.12.

The target model was the quantization-aware-trained Gemma 4 E4B instruction-tuned checkpoint \href{https://huggingface.co/google/gemma-4-E4B-it-qat-w4a16-ct}{\texttt{google/gemma-4-E4B-it-qat-w4a16-ct}}~\cite{team2026gemma}. The checkpoint stores four-bit weights with sixteen-bit activations in the compressed-tensors format and is intended for optimized inference with vLLM. The model has approximately 4.5 billion effective parameters, 42 decoder layers, and a vocabulary of approximately 262,000 tokens. The same quantized target checkpoint was used in both experimental conditions: it generated tokens directly in the autoregressive condition and supplied the verification distributions in the MTP condition.

The dedicated MTP drafter was \href{https://huggingface.co/google/gemma-4-E4B-it-assistant}{\texttt{google/gemma-4-E4B-it-assistant}}~\cite{team2026gemma}. This assistant checkpoint predicts speculative candidate tokens for subsequent verification by the target model. In the MTP condition, vLLM was configured to use the assistant checkpoint as a dedicated MTP speculator with two speculative tokens per step. The assistant was therefore an explicitly deployed auxiliary model rather than an implicit or internally derived prediction head.

The two decoding conditions were therefore:
\begin{itemize}
    \item \textbf{Autoregressive}: the quantized target model generated the output without a speculative model.
    \item \textbf{MTP}: the same target model verified token proposals produced by the dedicated MTP drafter model, with a speculative depth of \(\gamma=2\).
\end{itemize}

Apart from the speculative configuration, the conditions used common serving parameters. Both services used a maximum model length of 8,192 tokens, a maximum sequence count of one, a maximum of 2,048 batched tokens, BF16 KV-cache storage, chunked prefill, prefix caching, the same generation configuration, and the same chat template. Tool selection and the Gemma 4 reasoning and tool-call parsers were enabled in both conditions. GPU memory utilization was limited to 0.80 for the controlled benchmark services.

Each condition was exposed through a separate systemd service. Only the service corresponding to the active decoding mode was permitted to run during clean benchmark collection. The benchmark wrapper verified the expected service state and rejected execution when an NVIDIA profiling process was detected.

\subsection{Workloads}
\label{subsec:workloads}
The benchmark used three workload classes representing different output structures: plain-text generation, reasoning-intensive generation, and tool calling. Each class contained three fixed prompts, producing nine prompt conditions per decoding mode.

\paragraph{Plain-text generation.}
\label{para:plain-text}
The plain-text workloads requested short, constrained natural-language outputs. Output validity required exactly eight non-empty lines, with each line containing between six and twelve words. These prompts provided comparatively open-ended generation while preserving a machine-checkable structural requirement.
\paragraph{Reasoning-intensive generation.}
\label{para:reasoning-intensive}
The reasoning workloads required the model to derive a deterministic numerical result before producing a constrained final answer. The streaming response exposed reasoning content separately from the final answer, allowing the client to identify the transition between the reasoning and final-output phases. Output validity required the expected final-answer string to appear in the final response.
\paragraph{Tool calling.}
\label{para:tool-calling}
The tool-calling workloads supplied a fixed tool schema and required exactly one structured tool invocation. Output validity required the selected tool name to match the expected tool and the reconstructed JSON arguments to equal the expected argument object. Parallel tool calls were disabled.

The same prompt content, system instruction, chat template, request configuration, and validation rule were used for the normal and MTP conditions. Each request also contained a paired experiment identifier, allowing equivalent normal and MTP runs to be linked during analysis.

\subsection{Experimental Procedure} \label{subsec:experimental-procedure}
The primary experiment followed Protocol v1.2 (Appendix~\ref{app:protocol}), which was frozen before measurement and paired with a SHA-256 integrity record. The benchmark matrix comprised \begin{equation}
    2\ \text{mode}
    \times
    3\ \text{workload}
    \times
    3\ \text{prompts}
    \times
    20\ \text{repetitions}
    =
    360\ \text{measured requests}.
\end{equation}
This produced 180 requests per decoding mode and 20 matched repetitions per prompt. Before each mode-workload group, three unmeasured warm-up requests were issued and stored separately. The autoregressive experiment was completed and validated before the MTP service was started.

Requests were submitted sequentially through the local vLLM streaming API using the temperature, top-\(p\), seed, chat template, and 1,024-token output limit fixed by the protocol. Each request preserved its raw stream, reconstructed output, token usage, timestamps, validity result, and derived summary. The completed experiment contained 360 unique run records. The protocol, scripts, results, logs, and profiling artifacts were preserved in a checksum-verified archive.

\subsection{Measurements and Validity} \label{subsec:measurements-validity}
The benchmark client used a monotonic high-resolution clock to record six possible timestamps:
\begin{itemize}
    \item request start;
    \item first meaningful output;
    \item first reasoning content;
    \item first final-answer content;
    \item first tool-call fragment; and
    \item request end.
\end{itemize}
A meaningful output event contained reasoning content, final-answer content, or a tool-call fragment. Metadata-only streaming events were not treated as output.

The resulting measurements were time to first output (TTFO), end-to-end latency, time to first final output, reasoning-associated duration, final- or tool-associated duration, completion length, and output throughput. The phase durations are described as ``associated'' because they follow client-observed stream boundaries and do not attribute every operation in an interval exclusively to one semantic phase.

Output throughput was calculated as \begin{equation}
    \text{Output throughput}
    =
    \frac{N_{\mathrm{completion}}-1}
        {T_{\mathrm{end}}-T_{\mathrm{first\ output}}},
    \label{eq:output-throughput}
\end{equation}
where subtracting the first completion token aligns the token count with the measured post-first output interval.

Each request was evaluated for execution validity, semantic-boundary availability, and workload-specific output correctness. All 360 requests were execution-valid and phase-mapping-valid; 349 also passed output validation. The remaining eleven reasoning requests, consisting of six exact-answer mismatches and five length-limited completions, were retained rather than replaced.

Comparisons are performed first between matched prompts. Metrics that describe successful execution may use all execution-valid requests, whereas analyses requiring a correct, completed trajectory use only output-valid requests. End-to-end latency is interpreted alongside completion length and output throughput to account for response-length variation.

\subsection{Profiling Methodology} \label{subsec:profiling-methodology}
Profiling was conducted separately from the clean benchmark, and profiled request timings were not used as benchmark results. Matched captures were collected for \texttt{plain\_001}, \texttt{reasoning\_001}, and \texttt{tool\_001} under both autoregressive and MTP decoding.

\subsubsection{Nsight Systems}
Six Nsight Systems reports captured CUDA, NVTX, and operating-system runtime activity. Each capture used a temporary vLLM instance, a 30-second settling interval, three unmeasured workload-matched warm-ups, and one target request.

Client NVTX markers identified request start, first output, first reasoning content, first final answer or tool call, and request end. These markers isolate the target request from model initialization and warm-up activity. The matched traces are used to compare repeated GPU execution cadence, inter-group gaps, kernel composition, invocation counts, and aggregate kernel duration. Six corresponding SQLite exports support reproducible numerical analysis.

\subsubsection{PyTorch Profiler}
Six matched PyTorch Profiler captures were collected for the same conditions after three unprofiled warm-ups. Each condition preserved a worker trace, an AsyncLLM frontend trace, and a profiler summary. The worker traces provide higher-level operational context for repeated model execution and for matrix, attention, and speculative-path operations where the recorded labels permit unambiguous interpretation.

All twelve target requests used for Nsight Systems and PyTorch Profiler analysis passed phase-mapping and output validation.

\subsubsection{Relationship to the Analytical Cost Model}
The theoretical coefficient \(c\) in Equation~\ref{eq:theoretical-speedup} requires isolated draft and target step costs. Because the collected traces do not provide unambiguous boundaries for measuring these costs independently, this study does not estimate \(c\) directly.

Where requested execution groups can be identified consistently, the analysis instead considers the implementation-level ratio \begin{equation}
    r
    =
    \frac{T_{\mathrm{MTP\ group}}}{T_{\mathrm{AR\ group}}},
    \label{eq:implementation-cost-ratio}
\end{equation}
which represents the relative duration of the complete implemented execution groups. The empirical quantity \(r\) is not treated as equivalent to the theoretical draft-step coefficient \(c\).

Nsight Compute is used only if the Nsight Systems and PyTorch Profiler analyses identify a specific kernel requiring hardware-level characterization.

\section{Results}
\label{sec:results}

\subsection{Dataset Validity}
\label{subsec:dataset-validity}
The primary experiment collected all 360 planned requests, comprising 180 autoregressive and 180 MTP executions. Each decoding mode contained 20 repetitions of nine prompts distributed equally across the three workload classes. All run identifiers were unique, all rows used Protocol v1.2, and no timing or token metric was missing.

All 360 requests passed benchmark and phase-mapping validation. Of these, 349 also passed workload-specific output validation: 177 of 180 autoregressive requests and 172 of 180 MTP requests. The eleven retained output-invalid requests were confined to the reasoning-intensive workloads. Six \texttt{reasoning\_001} requests completed normally but did not contain the exact expected final-answer string, while five \texttt{reasoning\_002} requests reached the 1,024-token output limit and lacked the required final answer. These requests were preserved rather than replaced.

Application-level timing and throughput results below use all 360 benchmark-valid requests because every request completed under the controlled deployment and retained valid timing boundaries. Results that require a completed semantic trajectory use the 349 output-valid requests.

\subsection{RQ1: Application-Level Performance}
\label{subsec:rq1-performance}
Table~\ref{tab:rq1-workload-medians} reports workload-level medians across the 60 benchmark-valid requests in each workload and decoding mode. MTP improved sustained generation performance in all three workloads. Median output throughput increased from 97.46 to 192.80 tokens/s for plain text, from 96.61 to 204.70 tokens/s for reasoning-intensive requests, and from 97.26 to 204.98 tokens/s for tool calling.
\begin{table}[t]
    \centering
    \small
    \begin{tabular}{llrrrr}
    \toprule
    Workload & Mode & TTFO (ms) & E2E (ms) & TPS & Tokens \\
    \midrule
    Plain text & AR & 77.92 & 5107.05 & 97.46 & 491.0 \\
            & MTP & 69.40 & 2676.02 & 192.80 & 499.5 \\
    Reasoning-intensive & AR & 80.19 & 8373.85 & 96.61 & 802.0 \\
            & MTP & 69.22 & 4074.32 & 204.70 & 811.0 \\
    Tool calling & AR & 79.76 & 3369.30 & 97.26 & 321.0 \\
            & MTP & 70.06 & 1630.76 & 204.98 & 321.0 \\
    \bottomrule
    \end{tabular}
    \caption{Workload-level medians across benchmark-valid requests. E2E denotes end-to-end latency and TPS denotes output tokens per second. Each row contains 60 requests.}
    \label{tab:rq1-workload-medians}
\end{table}

Prompt-matched effects are reported in Table~\ref{tab:rq1-prompt-effects}. Throughput improved for every prompt, with speedups ranging from \(1.91\times\) to \(2.19\times\). The median prompt-level throughput speedup was \(1.95\times\) for plain text, \(2.11\times\) for reasoning-intensive, and \(2.11\times\) for tool calling. TTFO improved modestly, with reductions of 10.0--11.6\% for plain text, 12.6--14.2\% for reasoning-intensive, and 11.7--12.8\% for tool calling.
\begin{table}[t]
    \centering
    \small
    \begin{tabular}{lrrr}
    \toprule
    Prompt & TTFO reduction & E2E speedup & TPS speedup \\
    \midrule
    \texttt{plain\_001} & 11.3\% & \(2.27\times\) & \(1.91\times\) \\
    \texttt{plain\_002} & 11.6\% & \(1.91\times\) & \(1.95\times\) \\
    \texttt{plain\_003} & 10.0\% & \(2.04\times\) & \(2.05\times\) \\
    \texttt{reasoning\_001} & 12.9\% & \(2.17\times\) & \(2.19\times\) \\
    \texttt{reasoning\_002} & 12.6\% & \(1.94\times\) & \(2.11\times\) \\
    \texttt{reasoning\_003} & 14.2\% & \(2.04\times\) & \(2.06\times\) \\
    \texttt{tool\_001} & 12.2\% & \(2.08\times\) & \(2.12\times\) \\
    \texttt{tool\_002} & 11.7\% & \(2.04\times\) & \(2.09\times\) \\
    \texttt{tool\_003} & 12.8\% & \(2.07\times\) & \(2.11\times\) \\
    \bottomrule
    \end{tabular}
    \caption{Prompt-matched effects calculated from the median of 20 repetitions per mode. E2E speedup is \(\widetilde{T}_{\mathrm{AR}}/\widetilde{T}_{\mathrm{MTP}}\), and TPS speedup is \(\widetilde{R}_{MTP}/\widetilde{R}_{AR}\).}
    \label{tab:rq1-prompt-effects}
\end{table}

The tool-calling workload provided the most direct latency comparison because paired modes produced identical median completion lengths for all three prompts. MTP reduced prompt-level median E2E latency by factors of \(2.04\times\) to \(2.08\times\), while increasing output throughput by \(2.09\times\) to \(2.12\times\). The narrow range across the three prompts indicates that the effect was consistent within this structured workload.

Reasoning-intensive requests also showed substantial acceleration. Across its three prompts, median E2E speedups ranged from \(1.94\times\) to \(2.17\times\), while throughput speedups ranged from \(2.06\times\) to \(2.19\times\). The output-valid sensitivity analysis produced workload-level median completion lengths of 811 tokens under both modes, with median E2E latency decreasing from 8467.60 ms under autoregressive decoding to 4074.32 ms under MTP. Thus, the reasoning result was retained when analysis was restricted to correct and completed outputs.

Plain-text E2E latency requires greater caution because generated lengths varied between modes and repetitions despite identical prompts and fixed decoding settings. For example, \texttt{plain\_001} had median completion lengths of 486 tokens under autoregressive decoding and 390.5 tokens under MTP, contributing to its \(2.27\times\) E2E difference. Nevertheless, output throughput increased for every plain-text prompt, including \texttt{plain\_002}, where the median completion lengths were similar at 439 and 446 tokens. The plain-text throughput result therefore indicates a generation-rate improvement even where E2E latency was affected by trajectory length.
Semantic-phase-associated durations followed the same direction. Across the nine prompts, reasoning-associated duration improved by factors ranging from \(1.85\times\) to \(2.33\times\), while final- or tool-associated duration improved by \(2.26\times\) to \(2.68\times\). These intervals are interpreted as client-observed phase-associated durations rather than exclusive measurements of internal reasoning or tool-generation operations.

\paragraph{Answer to RQ1.}
Two-token MTP improved every measured prompt on the reported performance metrics. Its primary effect was sustained generation acceleration: output throughput increased by approximately \(1.91\times\) to \(2.19\times\), and prompt-level median E2E latency improved by \(1.91\times\) to \(2.27\times\). TTFO improved consistently but by a smaller 10.0--14.2\%, showing that the larger benefit occurred after output generation began. Tool calling produced the most stable comparisons, reasoning retained approximately twofold acceleration under an output-valid sensitivity analysis, and plain-text E2E results required completion-length-aware interpretation.

\subsection{RQ2: Acceptance and Theoretical Correspondence}
\label{subsec:rq2-acceptance}
The MTP service emitted speculative-decoding telemetry at approximately 10-second intervals. Each record reported mean acceptance length, accepted and drafted token counts, per-position acceptance rates, and average draft acceptance rate. Because several sequential requests could contribute to one logging interval, these records are treated as workload-associated aggregate windows rather than per-request measurements.

The telemetry windows were aligned with the request sequence recorded by the benchmark wrapper. Windows containing only warm-up requests or crossing workload transitions were excluded. Owing to differences in end-to-end workload execution time, this produced 17 clean windows for plain text, 24 for reasoning-intensive requests, and 12 for tool calling. Table~\ref{tab:rq2-acceptance} reports the median values within each workload.
\begin{table}[t]
    \centering
    \small
    \begin{tabular}{lrrrr}
    \toprule
    Workload & Windows & \makecell{Mean acceptance\\length} & \makecell{Avg. Draft\\acceptance rate} & \makecell{Position 1 , 2\\acceptance rates}\\
    \midrule
    Plain text & 17 & 2.370 & 68.7\% & 74.3\% , 62.8\% \\
    Reasoning intensive & 24 & 2.575 & 78.9\% & 84.5\% , 73.2\% \\
    Tool calling & 12 & 2.595 & 79.8\% & 83.3\% , 75.9\% \\
    \bottomrule
    \end{tabular}
    \caption{Median speculative-decoding telemetry across clean workload-associated logging windows. Mean acceptance length includes the target-generated token and therefore ranges from 1 to 3 tokens for the 2-token speculative configuration.}
    \label{tab:rq2-acceptance}
\end{table}

Acceptance decreased with speculative position in every workload. Plain-text windows recorded median acceptance rates of 74.3\% for the first draft position and 62.8\% for the second. The corresponding rates were 84.5\% and 73.2\% for reasoning-intensive requests, and 83.3\% and 75.9\% for tool calling. The second speculative candidate was therefore less likely to be accepted than the first, although it still contributed useful progress in all three workloads.

The observed mean acceptance lengths quantify the average token progress obtained per target verification iteration. Autoregressive decoding advances by one token per iteration, whereas the MTP configuration could advance by as many as three tokens: one target-produced token and up to two accepted draft tokens. Median mean acceptance lengths of 2.370, 2.575, and 2.595 therefore indicate substantial additional progress for plain text, reasoning-intensive, and tool calling requests, respectively.

The application-level throughput gains reported in Section~\ref{subsec:rq1-performance} were smaller than these acceptance lengths. Median prompt-level throughput speedups were \(1.95\times\), \(2.11\times\), and \(2.11\times\) for plain text, reasoning-intensive requests, and tool calling, respectively. Thus, greater accepted progress generally corresponded to greater throughput, but accepted tokens did not translate one-for-one into wall-clock speedup.

This difference is consistent with the analytical model: acceptance increases useful progress per verification iteration, while the dedicated drafter, expanded target verification, acceptance processing, and associated runtime operations add execution cost. Because the runtime telemetry does not isolate draft-step and target-step costs, the theoretical cost coefficient \(c\) cannot be estimated directly from these records. Consequently, the present comparison evaluates the directional correspondence between acceptance and realized speedup rather than claiming an exact prediction from acceptance length alone.

\paragraph{Answer to RQ2.} Two-token MTP provided substantial useful speculative progress in all three workloads, with median mean acceptance lengths ranging from 2.370 to 2.595 tokens per verification iteration. Reasoning-intensive and tool-calling workloads achieved higher draft acceptance than plain text and also produced slightly larger median prompt-level throughput speedups. However, realized speedup remained below mean accepted token progress, demonstrating that acceptance is necessary but insufficient to predict wall-clock acceleration without accounting for the cost of the implemented speculative path.

\subsection{RQ3: GPU Execution Analysis}
\label{subsec:rq3-gpu-execution}
To examine the GPU-side mechanism underlying the application-level performance differences, we analyzed one matched autoregressive (AR) and multi-token prediction (MTP) Nsight Systems capture for each workload: plain text, reasoning-intensive generation, and tool calling. The analysis used the target profiling request (\texttt{r100}) from each condition. Kernel and CUDA Graph records were extracted programmatically from the corresponding Nsight Systems SQLite exports.

The analysis window began at \texttt{FIRST\_OUTPUT} and ended at \texttt{RUN\_END}. This boundary excluded prompt processing, prefill, and first-output production while retaining both reasoning-associated generation and final-answer or tool-call-associated generation. In all captures, \texttt{FIRST\_REASONING} occurred within approximately 0.008--0.011 ms of \texttt{FIRST\_OUTPUT}; therefore, the selected window included effectively the complete post-first-output reasoning phase. The \texttt{FIRST\_FINAL} and \texttt{FIRST\_TOOL\_CALL} markers were used only as semantic and diagnostic boundaries and did not truncate the GPU analysis window.

\paragraph{Measurement definitions.}
A \emph{kernel launch} denotes one individually recorded invocation in \texttt{CUPTI\_ACTIVITY\_KIND\_KERNEL}. Kernel-launch counts therefore assign equal count weight to operations of very different duration, from microsecond-scale elementwise kernels to millisecond-scale matrix kernels. A \emph{selected repeating CUDA Graph} denotes the trace-local repeating graph used to delimit recurrent execution: Graph 520/GraphExec 521 for AR and Graph 313/GraphExec 314 for MTP. We define a \emph{recurring execution group} as the start-to-start interval between consecutive executions of the selected repeating CUDA Graph:
\begin{equation}
    T_{\mathrm{group},i}
    =
    t_{\mathrm{graph},i+1}
    -
    t_{\mathrm{graph},i}
\end{equation}

Graph identifiers are local to the respective traces and do not carry meaning outside the captured executions.

For each profiler request, kernel-launch count, executions of the selected repeating CUDA Graph, and summed individual-kernel duration were normalized by the exact number of completion tokens reported by that request:
\begin{equation}
    K_{\mathrm{token}}
    =
    \frac{N_{\mathrm{kernel}}}{N_{\mathrm{completion}}},
\end{equation}

\begin{equation}
    G_{\mathrm{token}}
    =
    \frac{N_{\mathrm{graph}}}{N_{\mathrm{completion}}},
\end{equation}
and
\begin{equation}
    T_{\mathrm{kernel/token}}
    =
    \frac{\sum_{j=1}^{N_{\mathrm{kernel}}}\left(t_{\mathrm{end},j}-t_{\mathrm{start},j}\right)}{N_{\mathrm{completion}}}
\end{equation}
Here, $K_{\mathrm{token}}$ is the number of individually recorded kernel launches per completion token, and $G_{\mathrm{token}}$ is the number of executions of the selected repeating CUDA Graph per completion token. The latter is distinct from the number of complete start-to-start recurring execution groups: $N$ graph executions provide $N-1$ measurable cadence intervals. Finally, $T_{\mathrm{kernel/token}}$ is the summed duration of individually recorded kernel intervals per completion token. It is neither GPU wall-clock time nor end-to-end request latency. CUDA Graph intervals were analyzed separately and were not added to their constituent kernel durations, since doing so would double-count overlapping execution.

\paragraph{Recurring execution structure.}
AR exhibited one dominant repeating graph topology across all three workloads. Graph 520/GraphExec 521 was repeatedly executed, followed by individually visible supporting operations and a dominant \texttt{gemv2T\_kernel\_val} invocation before the next graph replay. In contrast, MTP exhibited a more complex recurring structure. Two sequences of six short graph executions were followed by one execution of the larger Graph 313/GraphExec 314. The repeated MTP graph ordering was:
\begin{equation}
    \begin{split}
        &298 \rightarrow 301 \rightarrow 304 \rightarrow 307 \rightarrow 310 \rightarrow 1, \\
        &298 \rightarrow 301 \rightarrow 304 \rightarrow 307 \rightarrow 310 \rightarrow 1, \\
        &313.
    \end{split}
\end{equation}

The two short sequences are structurally consistent with the configured two-token speculative depth. However, the SQLite records establish only their execution ordering, not their semantic operator role. They are therefore described as \emph{auxiliary MTP graph sequences}, rather than being assigned confirmed drafter-pass semantics.

\begin{figure}[t]
    \centering
    \includegraphics[width=\linewidth]{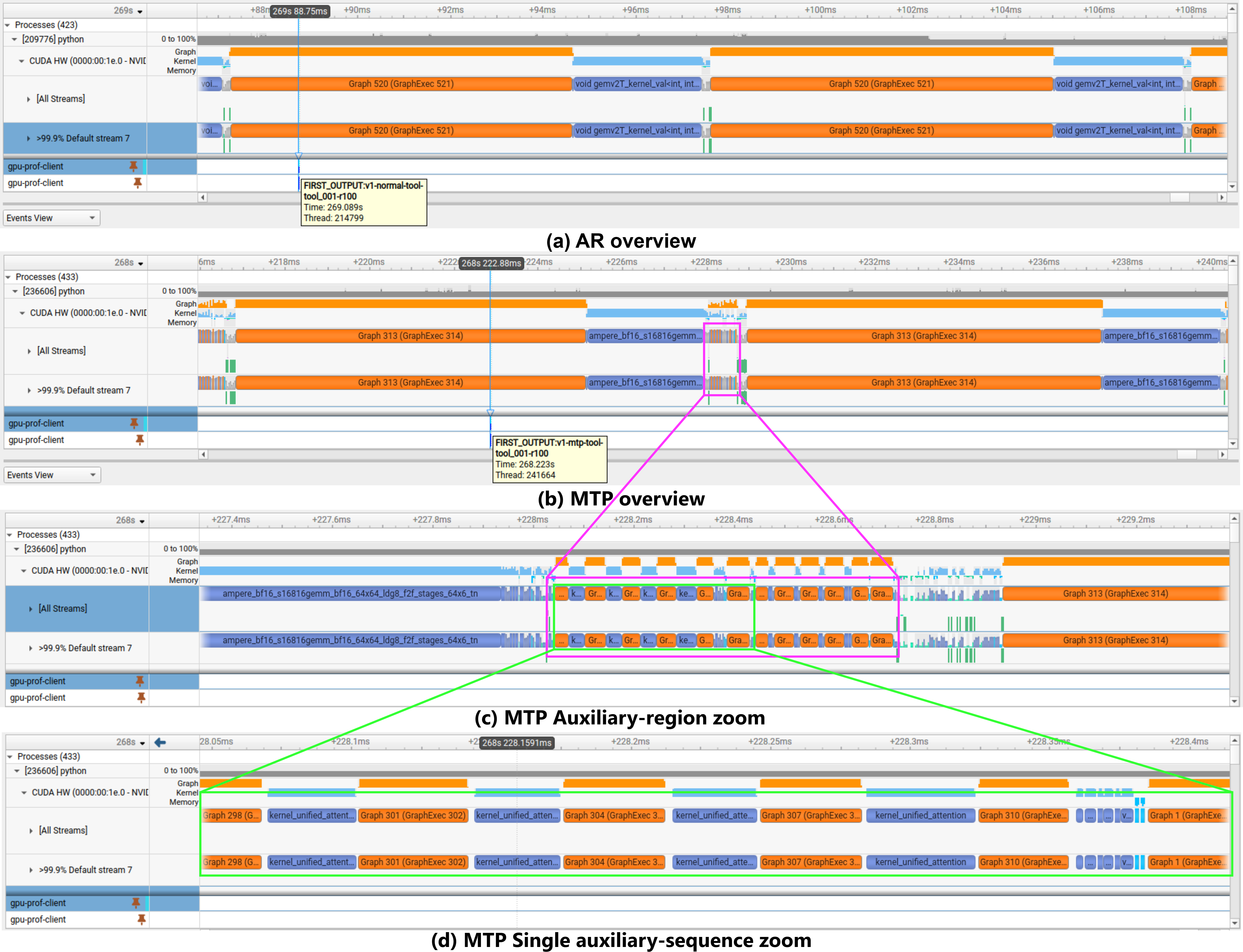}
    \caption{\textbf{Representative post-first-output GPU execution from the matched tool-calling captures.} Panel (a) shows consecutive executions of the AR repeating CUDA Graph, Graph 520/GraphExec 521, with the intervening \texttt{gemv2T\_kernel\_val}-dominated path. Panel (b) shows consecutive executions of the MTP repeating CUDA Graph, Graph 313/GraphExec 314, with the intervening BF16 GEMM and compact auxiliary graph and kernel activity. Panel (c) enlarges the compact MTP activity between consecutive Graph 313 executions and exposes two auxiliary graph sequences. Panel (d) enlarges one sequence, showing the ordering $298 \rightarrow 301 \rightarrow 304 \rightarrow 307 \rightarrow 310 \rightarrow 1$, interleaved with short supporting kernel activity. A recurring execution group was defined as the start-to-start interval between consecutive executions of the selected repeating CUDA Graph. The screenshots are illustrative; graph ordering, kernel identities, execution counts, durations, and cadence statistics were calculated from the Nsight Systems SQLite exports.}
    \label{fig:rq3-recurring-execution}
\end{figure}

Figure~\ref{fig:rq3-recurring-execution} illustrates this structural difference using the matched tool-calling captures. The figure is intentionally shown at a scale that preserves consecutive executions of the selected repeating CUDA Graph and one complete recurring execution group. Consequently, the short auxiliary MTP operations appear visually compressed. Their identities, counts, and durations were obtained from the SQLite records rather than estimated from the displayed widths.

\paragraph{Repeated-execution cadence.}
Table~\ref{tab:rq3-execution} reports the completion-token-normalized execution measurements. The median AR recurring-group cadence was highly stable across workloads, ranging from 10.363 to 10.383 ms. The corresponding MTP cadence ranged from 12.108 to 12.412 ms. We define the workload-specific cadence ratio as:
\begin{equation}
    r_{\mathrm{cadence}}
    =
    \frac{\widetilde{T}_{\mathrm{group,MTP}}}{\widetilde{T}_{\mathrm{group,AR}}},
\end{equation}
where $\widetilde{T}$ denotes the median interval between consecutive executions of the selected repeating CUDA Graph.

The resulting $r_{\mathrm{cadence}}$ values were 1.168 for plain text, 1.197 for reasoning-intensive generation, and 1.194 for tool calling. Thus, the median start-to-start cadence of an observed MTP recurring execution group was 16.8--19.7\% longer than that of the corresponding AR group. This longer cadence is consistent with the additional auxiliary graph sequences and supporting attention, gather, reduction, indexing, and rejection/sampling operations present in the MTP path.

\begin{table*}[t]
    \centering
    \resizebox{\textwidth}{!}{\begin{tabular}{llrrrrr}
    \toprule
    \textbf{Workload} & \textbf{Mode} & \makecell{\textbf{Completion}\\\textbf{tokens}} & \makecell{\textbf{Kernel launches}\\\textbf{per token}} & \makecell{\textbf{Individual-kernel}\\\textbf{ms per token}} & \makecell{\textbf{Repeating CUDA Graph}\\\textbf{executions per token}} & \makecell{\textbf{Median}\\\textbf{cadence (ms)}} \\
    \midrule
    Plain text & AR & 532 & 26.476 & 2.232 & 0.778 & 10.363 \\
    Plain text & MTP & 531 & 23.567 & 0.961 & 0.309 & 12.108 \\
    \addlinespace
    Reasoning-intensive & AR & 590 & 23.354 & 1.969 & 0.686 & 10.366 \\
    Reasoning-intensive & MTP & 588 & 22.854 & 0.945 & 0.299 & 12.412 \\
    \addlinespace
    Tool calling & AR & 321 & 22.271 & 1.881 & 0.654 & 10.383 \\
    Tool calling & MTP & 321 & 11.087 & 0.460 & 0.143 & 12.399 \\
    \bottomrule
    \end{tabular}}
    \caption{Post-first-output GPU execution measurements normalized by the exact completion-token count of each matched profiler request. Median cadence denotes the median start-to-start interval between consecutive executions of the selected repeating CUDA Graph. Kernel duration denotes summed individual-kernel duration and must not be interpreted as GPU wall-clock or end-to-end latency.}
    \label{tab:rq3-execution}
\end{table*}

Although the median MTP recurring-group cadence was longer, MTP required substantially fewer executions of the selected repeating CUDA Graph relative to generated output. Executions of the selected repeating CUDA Graph per token fell by 60.3\% for plain text, 56.4\% for reasoning-intensive generation, and 78.1\% for tool calling. The reduction in graph-execution frequency was substantially larger than the increase in recurring-group cadence. Consequently, summed individual-kernel duration per token fell by 57.0\%, 52.0\%, and 75.5\%, respectively.

\begin{table}[t]
    \centering
    \small
    \begin{tabular}{lrrrr}
    \toprule
        \textbf{Workload} & \makecell{\textbf{Launches}\\\textbf{per token}} & \makecell{\textbf{Kernel ms}\\\textbf{per token}} & \makecell{\textbf{Repeating CUDA Graph}\\\textbf{executions per token}} & $\boldsymbol{r_{\mathrm{cadence}}}$ \\
    \midrule
    Plain text & 11.0\% & 57.0\% & 60.3\% & 1.168 \\
    Reasoning-intensive & 2.1\% & 52.0\% & 56.4\% & 1.197 \\
    Tool calling & 50.2\% & 75.5\% & 78.1\% & 1.194 \\
    \bottomrule
    \end{tabular}
    \caption{Relative change from AR to MTP in the matched profiler captures. Positive values in the first three columns denote reductions under MTP.}
    \label{tab:rq3-relative}
\end{table}

The smaller change in kernel-launch density, especially for
reasoning-intensive generation, does not contradict the reduction in repeating CUDA Graph executions. A kernel launch is one operation, whereas a recurring execution group contains many operations. In the reasoning-intensive capture, AR recorded 13,779 kernel launches and 405 executions of Graph 520, yielding a capture-level ratio of approximately 34.0 kernel launches per selected graph execution. MTP recorded 13,438 kernel launches and 176 executions of Graph 313, yielding approximately 76.4 kernel launches per selected graph execution. The corresponding capture-level ratio was therefore approximately 2.25 times higher under MTP, consistent with the additional speculative-support path. Consequently, executions of the selected repeating CUDA Graph per token fell by 56.4\%, while total kernel launches per token fell by only 2.1\%. Most of the additional MTP operations were short relative to the dominant matrix kernel.

\paragraph{Kernel composition.}
The change in kernel composition was consistent across all workloads. AR execution was dominated by \texttt{gemv2T\_kernel\_val}, which accounted for approximately 97.06\% of summed individual-kernel duration in every AR capture. MTP execution was instead dominated by an \texttt{ampere\_bf16\_s16816gemm} kernel, which accounted for 88.57--89.52\% of summed individual-kernel duration.

\begin{table*}[t]
    \centering
    \resizebox{\linewidth}{!}{
    \begin{tabular}{llrrrr}
    \toprule
    \textbf{Workload} & \textbf{Mode and dominant kernel} & \textbf{Instances} & \makecell{\textbf{Total}\\\textbf{duration (ms)}} & \makecell{\textbf{Duration}\\\textbf{share}} & \makecell{\textbf{Median}\\\textbf{invocation ($\mu$s)}} \\
    \midrule
    Plain text & AR: \texttt{gemv2T\_kernel\_val} & 415 & 1152.511 & 97.07\% & 2777.066 \\
    Plain text & MTP: \texttt{ampere\_bf16\_s16816gemm...} & 164 & 456.571 & 89.52\% & 2784.122 \\
    \addlinespace
    Reasoning-intensive & AR: \texttt{gemv2T\_kernel\_val} & 406 & 1127.524 & 97.06\% & 2777.003 \\
    Reasoning-intensive & MTP: \texttt{ampere\_bf16\_s16816gemm...} & 177 & 492.757 & 88.69\% & 2784.137 \\
    \addlinespace
    Tool calling & AR: \texttt{gemv2T\_kernel\_val} & 211 & 585.987 & 97.06\% & 2776.906 \\
    Tool calling & MTP: \texttt{ampere\_bf16\_s16816gemm...} & 47 & 130.847 & 88.57\% & 2784.042 \\
    \bottomrule
    \end{tabular}}
    \caption{Dominant matrix-kernel composition in the matched post-first-output windows. The shortened kernel names identify the exact families reported by Nsight Systems; complete demangled names are retained in the accompanying artifacts.}
    \label{tab:rq3-dominant-kernels}
\end{table*}

The median dominant-kernel duration changed very little. AR GEMV invocations required approximately 2.777 ms, whereas MTP BF16 GEMM invocations required approximately 2.784 ms. The MTP dominant kernel was therefore approximately 0.25\% longer per invocation, rather than faster. The reduction instead arose from invocation frequency. Dominant matrix-kernel invocations fell from 415 GEMVs to 164 GEMMs for plain text, from 406 to 177 for reasoning-intensive generation, and from 211 to 47 for tool calling. These correspond to reductions of 60.5\%, 56.4\%, and 77.7\%, respectively. The dominant-kernel invocation reductions closely matched the workload-specific reductions in executions of the selected repeating CUDA Graph per token.

MTP also introduced operation families not observed in the matched AR windows. These included \texttt{kernel\_unified\_attention}, \texttt{vectorized\_gather\_kernel}, \texttt{reduce\_segments}, and a kernel matched by both rejection and sampling search patterns. The rejection and sampling matches referred to the same underlying kernel instances and were therefore counted once rather than added as separate operation sets. Table~\ref{tab:rq3-auxiliary-kernels} reports the observed launch counts and durations.

\begin{table*}[t]
    \centering
    \resizebox{\textwidth}{!}{
    \begin{tabular}{lrrrrrrrr}
    \toprule
    & \multicolumn{2}{c}{\textbf{Unified attention}} & \multicolumn{2}{c}{\textbf{Vectorized gather}} & \multicolumn{2}{c}{\textbf{Reduce segments}} & \multicolumn{2}{c}{\textbf{Rejection/sampling}} \\
    \cmidrule(lr){2-3}
    \cmidrule(lr){4-5}
    \cmidrule(lr){6-7}
    \cmidrule(lr){8-9}
    \textbf{Workload} &
    \textbf{Launches} & \textbf{ms} &
    \textbf{Launches} & \textbf{ms} &
    \textbf{Launches} & \textbf{ms} &
    \textbf{Launches} & \textbf{ms} \\
    \midrule
    Plain text & 1320 & 25.190 & 493 & 1.093 & 660 & 1.800 & 165 & 0.353 \\
    Reasoning-intensive & 1416 & 32.404 & 530 & 1.170 & 708 & 1.989 & 177 & 0.392 \\
    Tool calling & 376 & 8.823 & 140 & 0.309 & 188 & 0.518 & 47 & 0.102 \\
    \bottomrule
    \end{tabular}}
    \caption{MTP-specific auxiliary kernel activity in the post-first-output windows. No matching instances were observed in the corresponding AR windows.}
    \label{tab:rq3-auxiliary-kernels}
\end{table*}

The auxiliary launches explain why kernel-launch count alone understates the execution change. For example, reasoning-intensive MTP reduced executions of the selected repeating CUDA Graph per token by 56.4\%, yet kernel launches per token fell by only 2.1\%. The MTP capture contained 229 fewer dominant matrix-kernel invocations than the matched AR capture, decreasing from 406 AR GEMV invocations to 177 MTP GEMM invocations, but introduced 1,416 unified-attention launches, 530 vectorized-gather launches, 708 segmented-reduction launches, and 177 overlapping rejection/sampling launches. These support operations were individually short: unified attention contributed 32.404 ms, vectorized gather 1.170 ms, segmented reduction 1.989 ms, and rejection/sampling 0.392 ms across the entire reasoning-intensive MTP window. The expensive matrix-kernel reduction therefore dominated the added auxiliary execution, reducing summed individual-kernel duration per token by 52.0\%.

The same relationship appeared in the other workloads. For plain text, the MTP capture contained 251 fewer dominant matrix-kernel invocations than the matched AR capture, and the summed duration of the dominant matrix kernel was approximately 60.4\% lower. After accounting for the added MTP support kernels, total individual-kernel duration per token remained 57.0\% lower. For tool calling, the MTP capture contained 164 fewer dominant matrix-kernel invocations than the matched AR capture, and the summed duration of the dominant matrix kernel was 77.7\% lower. Total individual-kernel duration per token was 75.5\% lower.

\paragraph{Answer to RQ3.}
Across the three matched workloads, MTP changed repeated GPU execution in two related ways. First, compared with AR's frequently repeated, GEMV-dominated execution structure, MTP exhibited a more complex recurring path containing two auxiliary CUDA Graph sequences, Graph 313/GraphExec 314, BF16 GEMM, unified attention, vectorized gather, segmented reduction, and rejection/sampling activity. This additional work was accompanied by a 16.8--19.7\% longer cadence for the observed MTP recurring execution groups.

Second, MTP substantially reduced how frequently the selected repeating CUDA Graph and dominant matrix operation were executed relative to generated output. Repeating CUDA Graph executions per token fell by 56.4--78.1\%, and dominant matrix-kernel invocations fell by 56.4--77.7\%. Consequently, summed individual-kernel duration per token fell by 52.0--75.5\%, despite the longer median recurring-group cadence under MTP.

The observed application-level MTP benefit therefore did not correspond to a faster dominant matrix kernel or to the uniform elimination of small kernel launches. Instead, MTP added many short support operations but amortized this additional work by requiring far fewer executions of the selected repeating CUDA Graph and dominant matrix-kernel invocations per generated token. The mechanism was consistent across workloads, while its magnitude remained workload dependent. Tool calling exhibited the largest reduction, plain text was intermediate, and reasoning-intensive generation exhibited the smallest reduction.

\paragraph{Interpretation boundaries.} These profiler measurements characterize the execution mechanism rather than replacing the clean application-level latency results. A kernel launch is not a logical decoding step, and a recurring execution group is not a direct count of generated or accepted tokens. The count of selected repeating CUDA Graph executions therefore must not be used to infer speculative acceptance length. Similarly, the two auxiliary graph sequences are structurally consistent with two-token speculative execution but are not assigned confirmed drafter semantics from the Nsight Systems SQLite evidence alone.

Finally, summed individual-kernel duration excludes CPU-side processing, runtime scheduling gaps, synchronization outside the recorded kernel intervals, token streaming, response parsing, and post-GPU client activity. It must therefore not be reported as end-to-end speedup or GPU wall-clock time. The representative timeline in Figure~\ref{fig:rq3-recurring-execution} illustrates the observed structure, while all reported counts, durations, normalized values, and cadence statistics originate from the programmatic SQLite analysis.

\subsection{RQ4: Runtime Operations and Implementation Costs}
\label{subsec:rq4-runtime-costs}
To explain why speculative token progress did not translate proportionally into wall-clock acceleration, we analyzed the matched PyTorch Profiler worker traces for plain text, reasoning-intensive generation, and tool calling. Unlike the Nsight Systems captures used for RQ3, these worker traces did not contain the client-side NVTX markers. The RQ4 analysis therefore does not reconstruct the client-visible output phases. Instead, it isolated the repeated steady-state generation contexts recorded in the worker trace and examined their CPU-side runtime hierarchy and associated GPU-side execution spans.

\paragraph{Measurement scope and context identification.}
The dominant steady-state generation annotations were \nolinkurl{execute\_context\_0(0)\_generation\_1(1)} for AR and \nolinkurl{execute\_context\_0(0)\_generation\_1(3)} for MTP. Each logical generation context was recorded twice: once as a CPU-side \nolinkurl{user\_annotation} and once as a GPU-side \nolinkurl{gpu\_user\_annotation}. The two records were paired using their shared \texttt{External id}. All selected contexts formed exact one-to-one pairs across all workloads: 532 AR and 221 MTP pairs for plain text, 590 AR and 224 MTP pairs for reasoning-intensive generation, and 321 AR and 124 MTP pairs for tool calling.

The CPU-side annotation denotes the inclusive host-side execution range of the recorded generation context. The GPU-side annotation denotes the associated GPU execution span. The latter is not interpreted as summed kernel duration or GPU utilization because it can include multiple kernels, memory operations, and intervals between device activities. One-off context variants labelled with \texttt{generation\_0(0)} were excluded from the steady-state context distributions because they did not belong to the repeatedly observed \texttt{generation\_1(...)} population.

For mode $m$, observed completion-token progress per selected steady-state generation context was calculated as:
\begin{equation}
    P_m
    =
    \frac{N_{\mathrm{completion},m}}{N_{\mathrm{context},m}}.
\end{equation}
The selected GPU-context span per completion token was calculated as:
\begin{equation}
    C_{\mathrm{GPU/token},m}
    =
    \frac{\sum_{i=1}^{N_{\mathrm{context},m}}T_{\mathrm{GPU\ context},i}}
    {N_{\mathrm{completion},m}}.
\end{equation}
The corresponding AR-to-MTP context-span improvement was:
\begin{equation}
    S_{\mathrm{context}}
    =
    \frac{C_{\mathrm{GPU/token,AR}}}
    {C_{\mathrm{GPU/token,MTP}}}.
\end{equation}
This ratio describes the selected GPU-associated context spans in the instrumented worker traces. It is not an end-to-end speedup and does not replace the clean application-level results in RQ1.

\paragraph{Token progress and generation-context frequency.}
Table~\ref{tab:rq4-contexts} reports the context counts and execution-span measurements. In every AR capture, the number of selected steady-state contexts exactly equalled the completion-token count, giving one completion token per AR context. MTP required only 221 contexts for 531 plain-text tokens, 224 contexts for 588 reasoning-intensive tokens, and 124 contexts for 321 tool-calling tokens. The corresponding MTP progress was 2.403, 2.625, and 2.589 completion tokens per context.

These values describe observed output progress relative to the recorded worker contexts. They are not direct measurements of accepted speculative length because the PyTorch worker trace does not explicitly associate each context with the exact number of accepted tokens. Nevertheless, the cross-workload values are consistent with the multi-token progress measured independently in RQ2.

\begin{table*}[t]
    \centering
    \resizebox{\linewidth}{!}{
    \begin{tabular}{llrrrrrrr}
    \toprule
    \textbf{Workload} & \textbf{Mode} & \makecell{\textbf{Completion}\\\textbf{tokens}} & \textbf{Contexts} & \makecell{\textbf{Tokens per}\\\textbf{context}} & \makecell{\textbf{Median CPU}\\\textbf{span (ms)}} & \makecell{\textbf{Median GPU}\\\textbf{span (ms)}} & \makecell{\textbf{GPU span per}\\\textbf{token (ms)}} & $\boldsymbol{S_{\mathrm{context}}}$ \\
    \midrule
    Plain text & AR & 532 & 532 & 1.000 & 5.423 & 10.453 & 10.465 & \multirow{2}{*}{1.485} \\
    Plain text & MTP & 531 & 221 & 2.403 & 7.227 & 16.696 & 7.045 & \\
    \addlinespace
    Reasoning-intensive & AR & 590 & 590 & 1.000 & 5.388 & 10.463 & 10.468 & \multirow{2}{*}{1.650} \\
    Reasoning-intensive & MTP & 588 & 224 & 2.625 & 7.203 & 16.581 & 6.343 & \\
    \addlinespace
    Tool calling & AR & 321 & 321 & 1.000 & 5.611 & 10.468 & 10.505 & \multirow{2}{*}{1.590} \\
    Tool calling & MTP & 321 & 124 & 2.589 & 7.534 & 16.908 & 6.606 & \\
    \bottomrule
    \end{tabular}}
    \caption{Steady-state generation-context measurements from the matched PyTorch Profiler worker traces. CPU and GPU spans are durations of the paired \texttt{user\_annotation} and \texttt{gpu\_user\_annotation} ranges, respectively. GPU span per token is the sum of selected GPU-context spans divided by completion tokens. $S_{\mathrm{context}}$ is the AR-to-MTP ratio of this quantity and must not be interpreted as clean end-to-end speedup or summed kernel time.}
    \label{tab:rq4-contexts}
\end{table*}

MTP reduced the number of selected steady-state generation contexts by 58.5\% for plain text, 62.0\% for reasoning-intensive generation, and 61.4\% for tool calling. However, the median MTP context had a substantially longer execution span. Median GPU-context span increased from 10.453--10.468 ms under AR to 16.581--16.908 ms under MTP, corresponding to MTP/AR ratios of 1.597, 1.585, and 1.615. Median CPU-context span increased by similarly consistent factors of 1.333, 1.337, and 1.343.

The additional per-context cost therefore offset part of the speculative progress. Although MTP advanced 2.403--2.625 completion tokens per context, the selected GPU-context span per completion token improved by only 1.485--1.650$\times$. The instrumentation-run end-to-end ratios were 1.410$\times$ for plain text, 1.554$\times$ for reasoning-intensive generation, and 1.491$\times$ for tool calling. These values are lower than the clean application-level ratios reported in RQ1 and are used only to provide context for the profiled executions. Profiling overhead, request boundary work, and execution outside the selected steady-state contexts prevent the context-span ratios from being algebraically equivalent to whole-request latency ratios.

\begin{figure*}[t]
    \centering
    \includegraphics[width=1.0\linewidth]{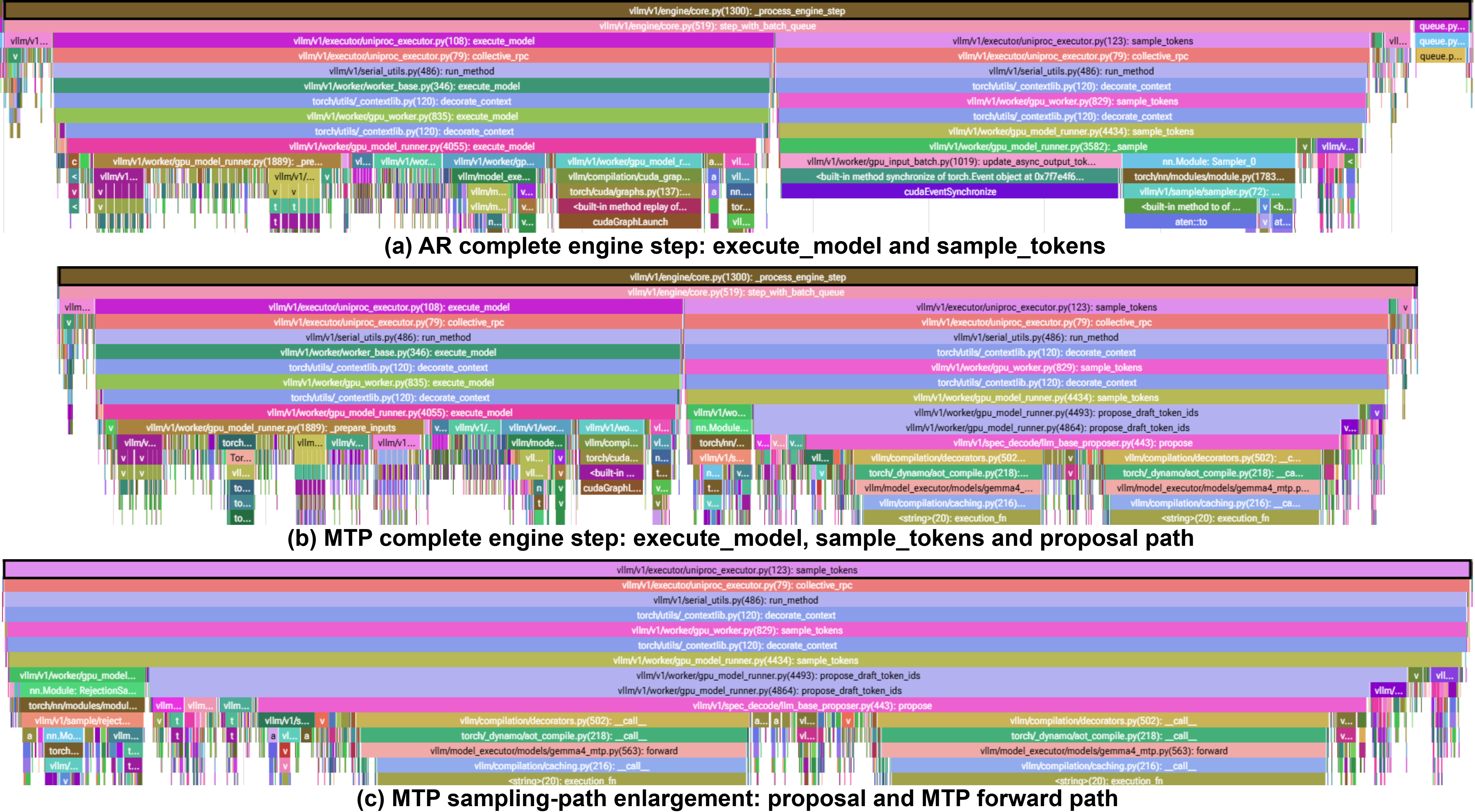}
    \caption{\textbf{Representative PyTorch Profiler runtime hierarchy from the matched tool-calling traces.} Panels (a) and (b) show one complete \texttt{\_process\_engine\_step} for AR and MTP, respectively, including the \texttt{execute\_model} and \texttt{sample\_tokens} paths. Panel (c) enlarges the MTP sampling path, in which \texttt{propose\_draft\_token\_ids} and \texttt{proposer.propose} are nested within \texttt{sample\_tokens}; two \texttt{gemma4\_mtp.py:forward} ranges are visible within the representative proposal subtree. The screenshots are illustrative; context pairing, function containment, call counts, and durations were calculated programmatically from the compressed PyTorch trace JSON artifacts.}
    \label{fig:rq4-runtime-hierarchy}
\end{figure*}

\paragraph{Explicit runtime operations.}
The PyTorch traces expose the higher-level vLLM runtime hierarchy that produced the additional MTP execution cost. In AR, the repeated runtime path contained \texttt{execute\_model} and \texttt{sample\_tokens}. In MTP, the sampling path additionally contained \texttt{propose\_draft\_token\_ids} and \texttt{proposer.propose}. Figure~\ref{fig:rq4-runtime-hierarchy} illustrates this runtime structure using the matched tool-calling traces. Programmatic timestamp containment confirmed the following hierarchy in every MTP workload:
\begin{equation}
    \texttt{sample\_tokens}
    \supset
    \texttt{propose\_draft\_token\_ids}
    \supset
    \texttt{proposer.propose}.
\end{equation}

All proposal calls satisfied this containment relationship. Plain text contained 444 of 444 proposal-ID calls inside \texttt{sample\_tokens} and 222 of 222 proposer calls inside a proposal-ID range. The corresponding counts were 450 of 450 and 225 of 225 for reasoning-intensive generation, and 250 of 250 and 125 of 125 for tool calling. No \texttt{propose\_draft\_token\_ids} or \texttt{proposer.propose} calls were observed in the matched AR worker traces.

\begin{table*}[t]
    \centering
    \resizebox{\linewidth}{!}{
    \begin{tabular}{llrrrrrr}
    \toprule
    \textbf{Workload} & \textbf{Mode} & \makecell{\textbf{\texttt{execute\_model}}\\\textbf{median (ms)}} & \makecell{\textbf{\texttt{sample\_tokens}}\\\textbf{median (ms)} }& \makecell{\textbf{Proposal-ID}\\\textbf{calls}} & \makecell{\textbf{Proposal-ID}\\\textbf{median (ms)}} & \makecell{\textbf{Proposer}\\\textbf{calls}} & \makecell{\textbf{Proposer}\\\textbf{median (ms)}} \\
    \midrule
    Plain text & AR & 5.333 & 4.128 & 0 & -- & 0 & -- \\
    Plain text & MTP & 7.110 & 9.140 & 444 & 7.857 & 222 & 6.957 \\
    \addlinespace
    Reasoning-intensive & AR & 5.294 & 4.205 & 0 & -- & 0 & -- \\
    Reasoning-intensive & MTP & 7.114 & 9.049 & 450 & 7.746 & 225 & 6.887 \\
    \addlinespace
    Tool calling & AR & 5.506 & 3.991 & 0 & -- & 0 & -- \\
    Tool calling & MTP & 7.405 & 9.154 & 250 & 7.854 & 125 & 6.967 \\
    \bottomrule
    \end{tabular}}
    \caption{Inclusive durations and call counts for selected vLLM runtime functions in the PyTorch worker traces. Proposal-ID denotes \texttt{propose\_draft\_token\_ids}, while proposer denotes \texttt{proposer.propose}. The displayed durations are inclusive Python range durations and must not be summed because the proposal ranges are nested within the sampling path. Aggregate call counts also include boundary executions outside the selected steady-state context set.}
    \label{tab:rq4-runtime-functions}
\end{table*}

Across the three workloads, median \texttt{execute\_model} duration was approximately 33.3--34.5\% longer under MTP. The larger difference appeared in \texttt{sample\_tokens}, whose median inclusive duration was approximately 2.15--2.29 times the AR duration. MTP also recorded approximately two \texttt{propose\_draft\_token\_ids} calls for every \texttt{proposer.propose} call. The proposal-related median durations were stable across workloads: 7.746--7.857 ms for \texttt{propose\_draft\_token\_ids} and 6.887--6.967 ms for \texttt{proposer.propose}.

These nested durations are not additive. In particular, the duration of \texttt{proposer.propose} contributes to the inclusive duration of \texttt{propose\_draft\_token\_ids}, which in turn contributes to the inclusive \texttt{sample\_tokens} range. The results therefore identify where the additional MTP runtime work occurs without claiming that the sum of these ranges is an independent latency decomposition.

\paragraph{Connection to GPU execution.}
The reduced number of steady-state MTP generation contexts observed in PyTorch Profiler is consistent with the lower frequency of repeating CUDA Graph executions measured independently in Nsight Systems. However, the two events are not treated as equivalent units: a PyTorch generation context is a high-level GPU-associated execution range that may contain multiple CUDA Graph executions, kernels, memory operations, and intervening activity.

RQ3 established that MTP executed fewer repeating CUDA Graphs per output token but introduced auxiliary graph sequences, expanded attention, gathering, reduction, sampling-related activity, and a BF16 GEMM-dominated execution path. The PyTorch traces provide the corresponding runtime-level interpretation. The additional device activity occurs beneath an explicit draft-proposal path within \texttt{sample\_tokens}, together with candidate selection and sequence-state processing. The MTP traces additionally contain top-$k$ gathering, sorting, vectorized gathering, segmented reduction, scatter, and index-update operations. These lower-level operations corroborate the additional sampling and proposal work, while the Python hierarchy establishes its semantic placement in the runtime.

The raw operator and kernel counts were not interpreted as independent latency contributions. CUDA Graph replay can change the visibility of operator ranges, and the same operation name can occur in different runtime contexts. RQ4 therefore uses the explicitly recorded runtime hierarchy, paired context spans, and completion-token normalization as its primary evidence. The kernel-level counts and durations remain part of RQ3.

\paragraph{Answer to RQ4.}
Across the three matched workloads, MTP reduced the number of steady-state generation contexts by 58.5--62.0\% and advanced 2.403--2.625 completion tokens per context. This speculative progress did not translate proportionally into wall-clock acceleration because each MTP context carried substantially more runtime and GPU-associated work. Median GPU-context span was 58.5--61.5\% longer under MTP, while median CPU-context span was 33.3--34.3\% longer.

PyTorch Profiler explicitly associated this additional work with a draft-proposal path absent from AR. All observed \texttt{propose\_draft\_token\_ids} ranges were nested within \texttt{sample\_tokens}, and all \texttt{proposer.propose} ranges were nested within proposal-ID processing. MTP \texttt{sample\_tokens} duration was approximately 2.15--2.29 times the AR duration, while \texttt{execute\_model} was approximately one-third longer. The resulting selected GPU-context span per completion token improved by 1.485--1.650$\times$, rather than by the full 2.403--2.625$\times$ completion-token progress observed per context.

The gap between speculative progress and realized acceleration therefore reflects the cost of producing and processing speculative candidates, not a failure to achieve multi-token progress. MTP reduced the number of repeated generation executions, but each execution incorporated proposal, expanded sampling, candidate-selection, tensor-manipulation, and state-update work. These per-context costs offset part of the useful token-progress advantage.

\paragraph{Interpretation boundaries.}
The PyTorch Profiler traces were instrumented executions and must not replace the clean benchmark results used for application-level performance claims. The selected GPU annotation spans are associated execution ranges rather than summed kernel durations or direct GPU-utilization measurements. Likewise, completion tokens per context describe aggregate output progress and must not be interpreted as an exact per-context speculative acceptance count.

\subsection{Selected-Kernel Analysis}
\label{subsec:selected-kernel-analysis}
The Nsight Systems analysis identified a recurring \texttt{gemv2T\_kernel\_val} kernel in AR and an \texttt{ampere\_bf16\_s16816gemm} kernel in MTP. Because these dominant kernels had nearly identical median invocation durations, one representative instance of each was inspected using Nsight Compute. Both instances were taken from the matched reasoning-intensive captures. This analysis provides supporting microarchitectural evidence and is not intended to represent all GEMV or GEMM invocations.
\begin{table}[t]
    \centering
    \small
    \begin{tabular}{lrr}
    \toprule
    \textbf{Metric} & \textbf{MTP GEMM} & \textbf{AR GEMV} \\
    \midrule
    Duration (ms) & 2.79 & 2.78 \\
    Memory throughput (GB/s) & 575.07 & 579.04 \\
    Maximum memory bandwidth (\%) & 95.93 & 96.59 \\
    Compute throughput (\%) & 22.06 & 28.89 \\
    L1/TEX hit rate (\%) & 0.06 & 7.63 \\
    L2 hit rate (\%) & 4.59 & 4.12 \\
    Executed IPC (instr./cycle) & 0.18 & 0.54 \\
    Issue slots busy (\%) & 4.57 & 13.58 \\
    Streaming Multiprocessors (SM) busy (\%) & 22.06 & 13.58 \\
    Achieved occupancy (\%) & 8.35 & 66.33 \\
    Active warps per SM & 4.01 & 31.84 \\
    Registers per thread & 102 & 64 \\
    Shared memory per block (KB) & 49.15 & 2.56 \\
    \bottomrule
    \end{tabular}
    \caption{Nsight Compute measurements for representative dominant kernels from the reasoning-intensive AR and MTP captures.}
    \label{tab:selected-kernel-analysis}
\end{table}

\begin{figure}[t]
    \centering
    \includegraphics[width=1.0\linewidth]{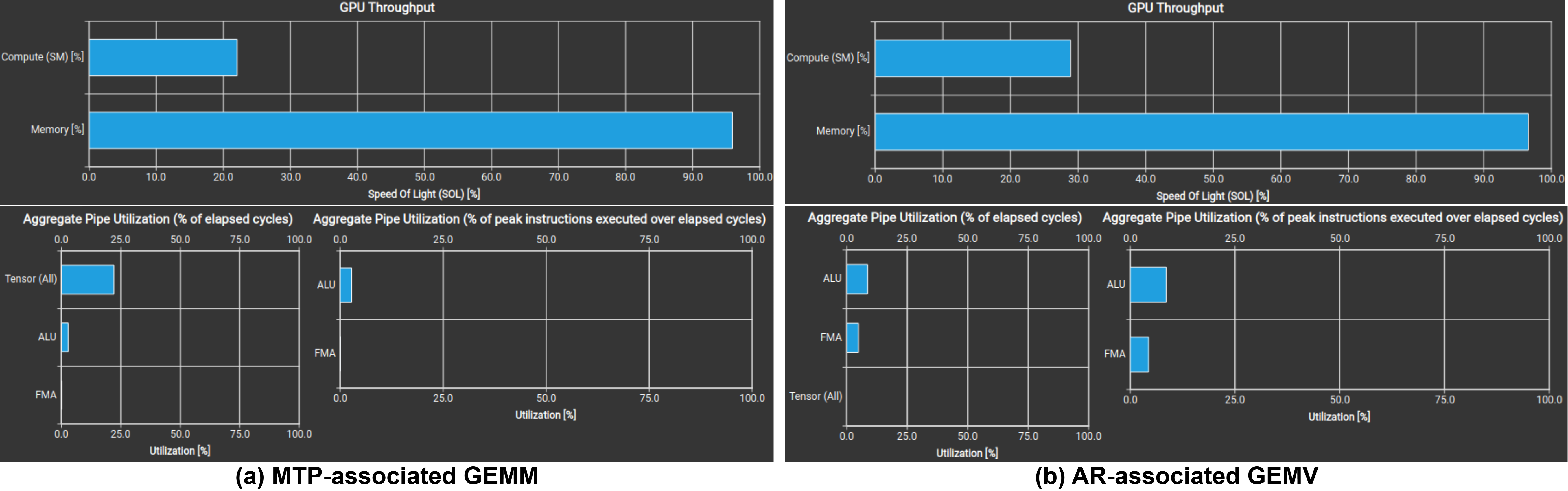}
    \caption{\textbf{Nsight Compute comparison of representative dominant kernels from the reasoning-intensive captures.} The MTP-associated GEMM (left) primarily used the Tensor pipeline, whereas the AR-associated GEMV (right) used conventional arithmetic pipelines. Both selected kernels approached the observed memory-bandwidth limit despite their different pipeline utilization.}
    \label{fig:selected-kernel-ncu}
\end{figure}

As shown in Table~\ref{tab:selected-kernel-analysis} and Figure~\ref{fig:selected-kernel-ncu}, both kernels operated near the A10G's nominal DRAM-bandwidth limit of 600~GB/s~\cite{nvidia2022a10g}. The MTP GEMM sustained 575.07~GB/s, or 95.93\% of maximum memory bandwidth, while the AR GEMV sustained 579.04~GB/s, or 96.59\%. Their low L2 hit rates further indicate that both selected instances depended heavily on device-memory traffic.  Therefore, the transition from GEMV-dominated AR execution to GEMM-dominated MTP execution did not correspond to a transition from memory-bound to compute-bound execution.

The kernels nevertheless used the GPU differently. The MTP GEMM primarily used the Tensor pipeline and recorded 22.06\% SM busy, but its resource footprint of 102 registers per thread and 49.15~KB of shared memory per block limited achieved occupancy to 8.35\%. The AR GEMV recorded lower SM busy at 13.58\%, but achieved 66.33\% occupancy with 64 registers per thread and 2.56~KB of shared memory per block while using load/store and conventional arithmetic pipelines.

These measurements reinforce the RQ3 result: MTP's advantage did not arise from a faster dominant kernel. The selected GEMM and GEMV instances had nearly identical durations and reached a similar memory-bandwidth ceiling. The workload-level improvement instead resulted from MTP requiring fewer dominant matrix-kernel and recurring group executions per generated token, despite introducing a more complex and individually costlier execution path.

\section{Discussion}
\label{sec:discussion}
\subsection{Useful Progress Versus Speculative Cost} \label{subsec:useful-progress-cost}
MTP accelerated generation because the reduction in repeated execution frequency outweighed its higher per-execution cost. Accepted proposals increased useful token progress, while proposal generation, expanded verification, sampling, and state updates made each MTP execution longer and more complex. The observed benefit therefore arose from amortizing this additional work across fewer recurring executions per generated token, not from making individual executions or dominant kernels faster.

\subsection{Analytical Model Versus Implemented Execution}
\label{subsec:theory-versus-implementation}
The results support the analytical model's central trade-off: speculation pays when additional token progress exceeds the cost of producing and verifying proposals. However, the theoretical draft-cost coefficient \(c\) could not be isolated from the traces. The measured cadence and context-span ratios instead describe the complete implemented path, including proposal, verification, sampling, scheduling, and device activity, and must not be treated as empirical estimates of \(c\).

\subsection{Practical Implications}
\label{subsec:practical-implications}
Speculative-decoding evaluation should consider accepted progress, execution frequency, and per-execution cost together. Acceptance rate alone cannot predict wall-clock speedup, while isolated kernel duration can conceal a substantial reduction in how often that kernel executes. For deployment and optimization, the relevant objective is therefore useful output progress relative to measured end-to-end latency, rather than acceptance or isolated kernel performance alone.

\section{Threats to Validity}
\label{sec:threats-to-validity}
The study evaluated one Gemma 4 target--drafter pairing, one NVIDIA A10G GPU, two-token speculation, single-request serving, and three prompts per workload, which limits generalization to other models, hardware, speculative depths, batch sizes, and concurrent deployments. Acceptance telemetry represented aggregate logging windows rather than individual requests, while the profiler analysis used one matched request per workload and mode. PyTorch contexts and CUDA Graph executions were observable implementation units rather than logical decoding steps, and the Nsight Compute analysis characterized only one selected GEMM and GEMV instance. Profiler results are therefore used as explanatory evidence, while application-level claims remain grounded in the separate clean benchmark.

\section{Future Work}
\label{sec:future-work}
This study deliberately prioritizes depth of performance analysis over breadth of configuration exploration. The speculative depth was intentionally fixed at \(\gamma\)=2 throughout the study. The objective was not to identify an optimal speculative depth or characterize speculative-depth scaling, but rather to explain the runtime and GPU-execution behavior of a deployed two-token MTP configuration. 

An important direction for future work is speculative-depth scaling, which could evaluate multiple \(\gamma\) values to examine how acceptance behavior, recurring GPU-execution cadence, proposal-processing overhead, and end-to-end performance evolve as speculative depth changes. Such a study could characterize the trade-off between speculative progress and the cost of proposal generation, verification, and acceptance processing.

\section{Conclusion}
\label{sec:conclusion}
In the evaluated deployment, two-token MTP increased output throughput by \(1.91\times\) to \(2.19\times\) across all prompts while producing smaller TTFO improvements of 10.0--14.2\%. Cross-layer profiling showed that this acceleration did not arise from a faster dominant kernel: MTP introduced a costlier proposal and verification path but required substantially fewer recurring GPU executions per generated token. The central result is therefore that MTP improved inference through amortization, with greater token progress more than compensating for the additional speculative-execution cost.

\printbibliography
\appendix
\section{Experimental Protocol}
\label{app:protocol}
The complete frozen experimental configuration is reproduced below. The protocol was finalized before primary measurement and preserved with its SHA-256 integrity record.
\begin{lstlisting}[
    style=yamlstyle,
    caption={Frozen experimental protocol, version 1.2.},
    label={lst:protocol-v1-2}
]
protocol:
  name: GPU Inference Profiling Study
  version: "1.2"
  status: frozen

research_question: >-
  How does MTP speculative decoding affect performance across plain-text,
  reasoning-intensive, and tool-calling workloads, and what execution behavior
  explains the differences?

deployments:
  normal: {service: prof-gemma4vllm.service, speculative_decoding: false}
  mtp: {service: prof-gemma4vllm-mtp.service, speculative_decoding: true, speculative_tokens: 2}

request:
  url: "http://127.0.0.1:8000/v1/chat/completions"
  model: "google/gemma-4-e4b"
  model_path: "<path_to_target_model>"
  drafter_path: "<path_to_drafter_model>"
  stream: true
  include_usage: true
  temperature: 0.0
  top_p: 1.0
  seed: 42
  benchmark_max_tokens: 1024
  profile_max_tokens: 1024
  parallel_tool_calls: false
  max_model_len: 8192
  max_num_seqs: 1
  gpu_memory_utilization: 0.80
  chunked_prefill: true
  max_num_batched_tokens: 2048
  kv_cache_dtype: "bfloat16"
  prefix_caching_enabled: true
  chat_template: "<path_to_target_model>/tool_chat_template_gemma4.jinja"
  chat_template_content_format: "openai"
  reasoning_parser: "gemma4"
  tool_call_parser: "gemma4"
  responses_api_store: false
  chat_template_kwargs:
    enable_thinking: true

phase_timing:
  clock: time.perf_counter_ns
  reasoning: delta.reasoning
  final_answer: delta.content
  tool_call: delta.tool_calls
  timestamps: [request_start_ns, first_output_ns, first_reasoning_ns, first_final_answer_ns, first_tool_call_ns, request_end_ns]
  nvtx_markers: [RUN_START, FIRST_OUTPUT, FIRST_REASONING, FIRST_FINAL, FIRST_TOOL_CALL, RUN_END]
  claim_limit: Profiler intervals are phase-associated; kernels are not inherently reasoning-specific.

tools:
  - type: function
    function:
      name: get_weather
      description: Get the weather forecast for a city and date.
      parameters:
        type: object
        properties:
          city: {type: string}
          date: {type: string, description: ISO date YYYY-MM-DD}
          unit: {type: string, enum: [celsius, fahrenheit]}
        required: [city, date, unit]
        additionalProperties: false
  - type: function
    function:
      name: get_local_time
      description: Get the current local time for a city.
      parameters:
        type: object
        properties: {city: {type: string}}
        required: [city]
        additionalProperties: false
  - type: function
    function:
      name: create_reminder
      description: Create a reminder.
      parameters:
        type: object
        properties:
          title: {type: string}
          datetime: {type: string, description: ISO 8601 date-time}
        required: [title, datetime]
        additionalProperties: false

workloads:
  plain_text:
    expected: reasoning followed by exactly eight poem lines
    prompts:
      - id: plain_001
        system: Follow the requested final format exactly. Add no title or commentary.
        user: Write exactly eight lines of free-verse poetry about a spacecraft travelling beyond the Solar System. Each line must contain six to twelve words.
      - id: plain_002
        system: Follow the requested final format exactly. Add no title or commentary.
        user: Write exactly eight lines of free-verse poetry about rain falling on a quiet city at night. Each line must contain six to twelve words.
      - id: plain_003
        system: Follow the requested final format exactly. Add no title or commentary.
        user: Write exactly eight lines of free-verse poetry about an old computer starting after many years. Each line must contain six to twelve words.

  reasoning_intensive:
    expected: reasoning followed by one strict FINAL line
    prompts:
      - id: reasoning_001
        system: Reason carefully. End with exactly one line in the required FINAL format.
        user: "A program is 80% perfectly parallelizable and 20% serial. Using Amdahl's law, calculate speedup with 8 processors and the maximum theoretical speedup. End exactly: FINAL: S8=<3 decimals>; Smax=<3 decimals>"
        expected_final: "FINAL: S8=3.333; Smax=5.000"
      - id: reasoning_002
        system: Reason carefully. End with exactly one line in the required FINAL format.
        user: "A 120-second workload is 75% perfectly parallelizable and 25% serial. Using Amdahl's law, calculate speedup and runtime with 6 processors. End exactly: FINAL: speedup=<3 decimals>; runtime=<3 decimals>s"
        expected_final: "FINAL: speedup=2.667; runtime=45.000s"
      - id: reasoning_003
        system: Reason carefully. End with exactly one line in the required FINAL format.
        user: "A program has serial fraction 0.10. Using Amdahl's law, find the minimum integer processors needed for speedup at least 5. End exactly: FINAL: processors=<integer>; speedup=<3 decimals>"
        expected_final: "FINAL: processors=9; speedup=5.000"

  tool_calling:
    expected: reasoning followed by exactly one get_weather call
    tool_choice: auto
    prompts:
      - {id: tool_001, system: Use exactly one available tool and do not answer directly., user: Get the weather forecast for Luxembourg City on 2026-10-15 in Celsius., expected_tool: get_weather, expected_arguments: {city: Luxembourg City, date: '2026-10-15', unit: celsius}}
      - {id: tool_002, system: Use exactly one available tool and do not answer directly., user: Get the weather forecast for Gurugram on 2026-11-20 in Fahrenheit., expected_tool: get_weather, expected_arguments: {city: Gurugram, date: '2026-11-20', unit: fahrenheit}}
      - {id: tool_003, system: Use exactly one available tool and do not answer directly., user: Get the weather forecast for Helsinki on 2026-12-05 in Celsius., expected_tool: get_weather, expected_arguments: {city: Helsinki, date: '2026-12-05', unit: celsius}}

cache:
  primary: controlled_cold
  method: Insert a pre-generated unique identifier at the beginning of each user message; reuse the identical prepared payload for paired normal and MTP runs.
  warm_cache_experiment: deferred

warmup:
  settle_after_health_seconds: 30
  minimum_per_workload: 3
  maximum_per_workload: 10
  require_two_consecutive_requests_without_jit_warning: true

repetitions:
  pilot: {prompts_per_workload: 2, repetitions_per_prompt: 5}
  primary: {prompts_per_workload: 3, valid_repetitions_per_prompt: 20}

metrics: [time_to_first_output_ms, time_to_first_final_output_ms, end_to_end_latency_ms, reasoning_associated_duration_ms, final_or_tool_associated_duration_ms, prompt_tokens, completion_tokens, output_tokens_per_second]

validity:
  benchmark_invalid_if: [JIT during measurement, wrong service or uncontrolled GPU work, HTTP or stream failure, profiler active during benchmark, cache condition not achieved]
  phase_mapping_invalid_if: [phase boundaries unavailable, required phase timestamp missing]
  output_invalid_if: [finish_reason is length, required final output absent, tool call or arguments mismatch]
  preserve_invalid_runs: true

artifacts:
  results: results.csv
  per_run: [runs/<run_id>/raw.jsonl, runs/<run_id>/summary.json]
  profiles: [profiles/torch, profiles/nsys, profiles/ncu]
  run_id: v1-<mode>-<workload>-<prompt_id>-r<repetition>
\end{lstlisting}

\end{document}